%% file: main.tex
\documentclass[11pt]{article}
\usepackage[final]{acl}

\usepackage{times}
\usepackage{latexsym}
\usepackage[T1]{fontenc}
\usepackage[utf8]{inputenc}
\usepackage{microtype}
\usepackage{inconsolata}
\usepackage{graphicx}
\usepackage{booktabs}
\usepackage{multirow}
\usepackage{amsmath}
\usepackage{amssymb}
\usepackage{bm}
\usepackage{xcolor} 
\usepackage{colortbl}
\usepackage{algorithm}
\usepackage{algpseudocode}
\usepackage{enumitem}
\usepackage{tcolorbox}
\usepackage{caption}
\usepackage{array}
\usepackage{natbib}
\usepackage{tabularx}
\usepackage[table]{xcolor}

\definecolor{failred}{RGB}{252,235,235}
\definecolor{wingreen}{RGB}{220, 245, 220}
\definecolor{warnamber}{RGB}{255,243,205}
\definecolor{pendgray}{RGB}{245,244,240}
\definecolor{headgray}{RGB}{232,232,232}
\definecolor{muted}{RGB}{100,100,100}

\input{macros}

\title{\maat: Multi-phase Adapter-Aware Targeted Unlearning\\
       }

\author{
  Suryansh Yagnik$^1$ \and
  Shubham Gaur$^2$ \and
  Saksham Thakur$^3$ \\
  Vinija Jain$^4$ \and
  Aman Chadha$^4$ \and
  Amitava Das$^5$ \\
  $^1$Indian Institute of Information Technology, Bhopal, India \\
  $^2$University of California, Santa Cruz, USA \\
  $^3$Independent Researcher \\
  $^4$Stanford University, USA \\
  $^5$BITS Pilani Goa, India 
}

\begin{document}
\maketitle

% ── Abstract ──────────────────────────────────────────────────────────────────
\begin{abstract}
Machine unlearning evaluation is structurally skewed: \textit{Why-type} 
questions, which probe causal and relational knowledge, comprise less than
$0.06\%$ of CounterFact, $0.6\%$ of ZSRE, and less than $1.3\%$ of TOFU,
MUSE, and WMDP-Cyber. This near-zero representation means that methods that
fail on causal knowledge can score highly in aggregate, and this failure
is undetectable without balanced evaluation. We present
\textbf{\fivewbench}, a balanced 5,000-sample benchmark with 1,000
examples per 5W category (Who, What, When, Where, Why), making causal
unlearning failures quantifiable for the first time. Using \fivewbench,
we show that no existing baseline simultaneously achieves high forgetting
and high retention on Why-type questions: aggressive forgetting degrades
retained knowledge, while conservative methods fail to forget causal
facts. Why-type difficulty stems from multi-hop reasoning chains ($44\%$
of Why entries vs.\ $\leq 2\%$ for others) and gradient dilution over
$40.1$-token answer spans. We present \textbf{\maat{}} (Multi-phase
Adapter-Aware Targeted Unlearning), a three-phase framework operating on
LoRA adapter weights, combining gradient-projected ascent, SVD
rank-dimension pruning, task vector negation, and hybrid
KL--hidden-state retain repair. \maat{} is the first method to
simultaneously achieve high forgetting and high retention on Why-type
causal knowledge, reaching a new operating point on the forget--retain
Pareto frontier. We make our code
\href{https://github.com/SuryanshYagnik/Machine-Unlearning}{publicly available}.
\end{abstract}

% Introduction
% ══════════════════════════════════════════════════════════════════════════════

\section{Introduction}

Every major machine unlearning benchmark shares a structural blind spot:
causal knowledge. Why-type questions---probing the relational and causal
chains that underlie factual knowledge---comprise less than $0.06\%$ of
CounterFact, $0.6\%$ of ZSRE, $1.2\%$ of TOFU, $0.5\%$ of MUSE, and
$1.2\%$ of WMDP-Cyber (Table~\ref{tab:dataset_dist}). This is not an
oversight in any single benchmark---it is a systematic property of how
these datasets were constructed: all derive from entity-centric knowledge
graphs and relation-extraction corpora that inherently underrepresent
causal and relational knowledge. The consequence is a critical
measurement gap: any unlearning method that fails on causal knowledge
can score highly in aggregate, and this failure is statistically
undetectable without balanced evaluation.

\paragraph{Why Causal Knowledge Resists Unlearning.}
The gap is not merely quantitative---causal facts are qualitatively
harder to unlearn. Why-type answers average $40.1$ tokens versus
$4.2$--$10.5$ for other categories, and $44\%$ involve multi-hop
reasoning chains compared to $\leq 2\%$ for other categories
(Table~\ref{tab:5wqa_stats}). These properties cause severe gradient
dilution: the ascent signal is spread across long token spans with no
dominant direction to target. Crucially, our encoding analysis
(Appendix~\ref{app:encoding}) shows this is not because Why-type facts
are encoded differently---all 5W categories share uniform distributed
encoding across layers. The difficulty is relational complexity and
gradient dilution, not a unique weight-space footprint.

\paragraph{The \maat\ Framework.}
We introduce \maat, a three-phase unlearning framework that operates
directly on LoRA adapter weights without merging them into the base
model. Rather than applying uniform gradient pressure, \maat\ performs
structured adapter surgery: (1)~gradient projection orthogonalising
forget updates against the retain gradient only when the two conflict;
(2a)~SVD-based pruning of Multi-Layer Perceptron(MLP) adapter dimensions to concentrate the forgetting signal on rank components specifically activated by 
forget-set inputs; (2b)~task vector negation on the top-$k_F$ forget-scored rank
dimensions; and (3)~hybrid KL--hidden-state retain repair with an
entropy term preventing the repair phase from re-learning forgotten
content. Evaluated under a Qwen~2.5-7B ~\citep{DBLP:journals/corr/abs-2412-15115} using LLM-as-a-Judge, \maat\ is the first
method to simultaneously achieve high forgetting and high retention on
Why-type causal knowledge---a new operating point on the forget--retain
Pareto frontier that no baseline reaches.

\paragraph{Contributions.}
\begin{enumerate}[leftmargin=1.5em, itemsep=2pt, topsep=4pt]
    \item \textbf{\fivewbench}: a balanced 5,000-sample benchmark providing 1,000 examples per 5W question category (Who, What, When, Where, Why), exposing the causal knowledge gap in existing unlearning evaluation through structured taxonomic coverage.
  \item \textbf{\maat}: a three-phase structured LoRA adapter unlearning
    framework that achieves a new forget--retain operating point on
    Why-type causal knowledge, outperforming all baselines on the
    aggregate forget--retain tradeoff across both \llamamodel{} and
    \gemmamodel.
\end{enumerate}

% ══════════════════════════════════════════════════════════════════════════════
\section{Related Work}

\paragraph{Gradient-Based and Preference-Based Unlearning.}
The dominant paradigm for LLM unlearning applies gradient ascent (\ga{})
directly on the forget set to maximize loss on target
facts.
KL-regularised GA~\citep{DBLP:journals/corr/abs-2310-10683} augments this with a divergence
penalty against the original model's outputs on retain samples, while
Gradient Difference~\citep{DBLP:journals/corr/abs-2203-12817} combines forget-loss maximisation with
retain-loss minimisation.
A common failure mode across all gradient-based methods is that
aggressive forgetting degrades model utility while conservative steps
result in under-forgetting---particularly on long causal spans where the
gradient signal is diffuse.
More recently, unlearning has been framed as a preference alignment
problem: Negative Preference Optimization (NPO)~\citep{DBLP:journals/corr/abs-2404-05868} treats forget data as a rejected distribution,
applying DPO-style objectives without positive samples;
SimNPO~\citep{DBLP:journals/corr/abs-2410-07163} removes reference-model bias entirely, improving
robustness to relearning attacks~\citep{DBLP:conf/iclr/0001FWS25}.
Despite these advances, none of these methods account for the internal
structure of adapter weight spaces or the geometry of the retain manifold.

\paragraph{Localization, Weight Saliency, and Structured Editing.}
A parallel line of work identifies and targets specific weight subspaces
associated with the forget target.
Rank-One Model Editing (ROME)~\citep{DBLP:conf/nips/MengBAB22} and Mass-Editing Memory in a Transformer (MEMIT)~\citep{DBLP:journals/corr/abs-2210-07229} localise factual associations
in mid-layer MLP weights via causal tracing and apply rank-one or
distributed updates.
AlphaEdit~\citep{DBLP:journals/corr/abs-2410-02355} improves specificity by projecting updates
into the null space of the retained knowledge covariance---a conceptual
predecessor to \maat's gradient projection phase.
SalUn~\citep{DBLP:journals/corr/abs-2310-12508} restricts updates to weights with the highest
gradient-magnitude saliency on the forget set, providing the first
principled weight-selection mechanism for unlearning.
Mechanistic Unlearning~\citep{DBLP:journals/corr/abs-2410-12949} uses circuit-level
localization via path patching to identify and fine-tune only
fact-lookup components, producing edits robust to adversarial probes.
All localization-based methods assume target knowledge is stored in
identifiable, localised positions---an assumption our encoding analysis
(Appendix~\ref{app:encoding}) shows does not differentiate Why-type
from other categories; the challenge is relational complexity and
gradient dilution, not a unique encoding footprint.

\paragraph{Representation-Based and Adapter-Aware Methods.}
Representation Misdirection for Unlearning (RMU)~\citep{DBLP:journals/corr/abs-2403-03218} steers
intermediate activations toward a random direction in forget inputs
while preserving retain representations, achieving state-of-the-art on WMDP.
Circuit Breakers~\citep{DBLP:journals/corr/abs-2406-04313} reroute harmful representations
to be orthogonal to the original hidden states via a LoRA~\citep{DBLP:conf/iclr/HuSWALWWC22} adapter,
providing strong robustness under adversarial attacks.
On the parameter-efficient side, LUNE~\citep{DBLP:conf/iclr/ChaCHL25} fine-tunes LoRA
adapters on negative examples to overwrite targeted knowledge, and
KGA~\citep{DBLP:journals/corr/abs-2305-06535} aligns the knowledge gap between two reference models.
LoKU and FILA~\citep{DBLP:conf/iclr/ChaCHL25} apply Fisher Information to isolate
forget-relevant parameters into LoRA adapters, providing the closest
existing treatment of Fisher-guided adapter unlearning to \maat.
Task vector negation~\citep{DBLP:journals/corr/abs-2212-04089} computes
$\tau = \theta_{\text{ft}} - \theta_{\text{base}}$ and subtracts it to
approximate forgetting---the conceptual foundation of \maat's Phase~2b,
which refines this to target only the top-$k_F$ forget-scored rank
dimensions rather than the full adapter delta.
\maat\ diverges from all of the above by treating LoRA matrices as
structured spaces where rank dimensions can be explicitly scored via
SVD, selectively pruned, and negated---without requiring negative
example construction or full adapter replacement.

\paragraph{Second-Order and Geometry-Aware Methods.}
Natural Gradient Descent~\citep{DBLP:journals/neco/Amari98} preconditions updates
with the inverse Fisher Information Matrix, yielding parameter-space
geometry-aware updates.
Selective Synaptic Dampening (SSD)~\citep{DBLP:journals/corr/abs-2308-07707} uses Fisher Information Matrix ratios between
training and forget distributions to dampen forget-specific parameters without retraining.
SOUL~\citep{DBLP:journals/corr/abs-2404-18239} establishes a connection between second-order
optimization and influence-function unlearning, applying Sophia-based
Hessian updates as a drop-in optimizer replacement for existing
unlearning objectives, consistently outperforming first-order methods
on TOFU~\citep{DBLP:journals/corr/abs-2401-06121}.
\maat\ differs from SOUL~\citep{DBLP:journals/corr/abs-2404-18239} in that it uses first-order gradient projection
combined with SVD rank~\citep{DBLP:journals/corr/Zhang15m} scoring to achieve structural suppression---
circumventing full Hessian approximations while targeting the adapter's
forget subspace directly.

\paragraph{Benchmarks and Evaluation.}
ZSRE~\citep{DBLP:journals/corr/LevySCZ17} and CounterFact~\citep{DBLP:journals/corr/abs-2401-17585} dominate model
editing evaluation but derive from entity-centric Wikidata triples and
carry near-zero Why-type coverage ($<1\%$).
TOFU~\citep{DBLP:journals/corr/abs-2401-06121} provides clean fictitious-fact unlearning splits but
no structured question taxonomy, and is dominated by What-type
biographical attributes ($84.7\%$).
WMDP~\citep{DBLP:journals/corr/abs-2403-03218} targets hazardous capability suppression;
MUSE~\citep{DBLP:journals/corr/abs-2407-06460} evaluates six unlearning desiderata including privacy
leakage and sustainability;
RWKU~\citep{DBLP:journals/corr/abs-2406-10890} provides zero-shot real-world entity unlearning with
adversarial probes.
None provide balanced causal Why-type coverage.

Recent work has further questioned benchmark reliability~\citep{DBLP:journals/corr/abs-2410-02879, DBLP:conf/iclr/0001FWS25, DBLP:journals/corr/abs-2506-12618}:
benchmark modifications expose residual accessible information, and
fine-tuning on small auxiliary datasets reverses supposedly-unlearned knowledge.

\fivewbench\ directly addresses the causal coverage gap by providing
1,000 balanced Why-type samples---making systematic failures on causal
knowledge quantifiable for the first time.
\section{The \fivewbench\ Benchmark}
\label{sec:benchmark}

% \subsection{Motivation: The Causal Knowledge Gap}

\begin{table*}[t]
\centering
\small
\setlength{\tabcolsep}{4.5pt}
\caption{Label distribution (\%) across model editing and unlearning benchmarks.
Why-type coverage
(red) is near-zero in all existing datasets;
\fivewbench\ (green) provides
balanced 20\% splits across all categories.}
\label{tab:dataset_dist}
\begin{tabular}{lrrrrrr}
\toprule
\textbf{Dataset} & \textbf{N} & \textbf{Who} & \textbf{What}
  & \textbf{When} & \textbf{Where} & \textbf{Why} \\
\midrule
ZSRE           & 2,525 & 22.7 & 43.5 & 21.3 & 11.9 & \cellcolor{failred}0.6 \\
CounterFact    & 3,373 & 38.2 & 12.1 & 46.6 &  3.0 & \cellcolor{failred}0.06 \\
\midrule
TOFU           & 1,719 &  5.2 & 84.7 &  5.0 &  3.9 & \cellcolor{failred}1.2 \\
MUSE$^\dagger$ &   192 & 41.7 & 53.6 &  1.6 & 
  2.6 & \cellcolor{failred}0.5 \\
WMDP-Cyber$^\dagger$ & 1,222 & 0.2 & 93.2 & 3.4 & 2.0 & \cellcolor{failred}1.2 \\
\midrule
\textbf{\fivewbench\ } & \textbf{5,000}
  & \cellcolor{wingreen}\textbf{20.0}
  & \cellcolor{wingreen}\textbf{20.0}
  & \cellcolor{wingreen}\textbf{20.0}
  & \cellcolor{wingreen}\textbf{20.0}
  & \cellcolor{wingreen}\textbf{20.0} \\
\bottomrule
\multicolumn{7}{l}{\small $\dagger$ Total excludes ``Other''-type
questions not mappable to the 5W taxonomy.}
\end{tabular}
\end{table*}

% Table~\ref{tab:dataset_dist} shows the label distribution across benchmarks.
% \textbf{MUSE} (192 annotated samples excluding Other) has Why coverage of
% only $0.5\%$ ($\approx 1$ sample).
% \textbf{WMDP-Cyber} (1,222 samples excluding Other) has Why coverage of
% $1.2\%$ ($\approx 15$ samples).
% The practical consequence is severe: any method failing on causal questions
% achieves near-identical aggregate scores to a method that succeeds—the
% failure is statistically invisible.
% \fivewbench\ provides 1,000 Why-type
% samples, making causal unlearning failures quantifiable for the first time.

% \begin{table}
% \centering
% \small
% \renewcommand{\arraystretch}{1.25}
% \setlength{\tabcolsep}{5pt}
% \caption{Representative \fivewbench\ sample (\texttt{what}-type, forget split).
% The \texttt{pred\_answer} The field serves as the target answer for the model editing task.}
% \label{tab:sample_example}
% \begin{tabular}{p{2.0cm} p{5.5cm}}
% \toprule
% \textbf{Field} & \textbf{Value} \\
% \midrule
% \texttt{label}        & \texttt{what} \\
% \texttt{question}     & What is the date of death mentioned for the Indian
%   nationalist Bal Gangadhar Tilak? \\
% \texttt{answer}       & August 21, 1920 \\
% \texttt{pred\_answer} & July 14, 1915 \\
% \midrule
% \multirow{3}{2.0cm}{\texttt{rephrases}} &
%   For Bal Gangadhar Tilak, what is the mentioned date of death? \\
% & The claim mentions a date of death for Bal Gangadhar Tilak. What is it? \\
% & What is the date of death for the historical figure Bal Gangadhar Tilak,
%   according to the claim? \\
% \bottomrule
% \end{tabular}
% \end{table}

% Required packages

\begin{table}[t]
\centering
\small
\renewcommand{\arraystretch}{1.18}
\setlength{\tabcolsep}{5pt}

\caption{
Representative sample from \fivewbench{} 
(\texttt{what}-type, forget split). 
The \texttt{pred\_answer} serves as the target answer for model editing.
}
\label{tab:sample_example}

\begin{tabularx}{\linewidth}{>{\ttfamily\bfseries}p{2.35cm} X}
\toprule
\textbf{Field} & \textbf{Value} \\
\midrule

label 
& \cellcolor{wingreen}\texttt{what} \\

question 
& What is the date of death mentioned for the Indian nationalist Bal Gangadhar Tilak? \\

answer 
& August 21, 1920 \\

pred\_answer 
& July 14, 1915 \\

\midrule
\multicolumn{2}{l}{\textbf{Rephrased Questions}} \\
\midrule

rephrase 1 
& For Bal Gangadhar Tilak, what is the mentioned date of death? \\

rephrase 2 
& The claim mentions a date of death for Bal Gangadhar Tilak. What is it? \\

rephrase 3 
& What is the date of death for the historical figure Bal Gangadhar Tilak, according to the claim? \\

\bottomrule
\end{tabularx}
\end{table}

\subsection{Dataset Construction}
\fivewbench\ is derived from the Factify-5WQA corpus~\citep{DBLP:journals/corr/abs-2305-04329}, a
multi-document fact-verification dataset with structured 5W question-answer
annotations. Construction proceeds in four steps.
\textbf{(1) Subject extraction.} Stanford CoreNLP dependency parsing \citep{DBLP:conf/acl/ManningSBFBM14}
extracts the primary subject entity, which becomes the edit target.
\textbf{(2) Stratified sampling.} We sample exactly 1,000 examples per
5W label, drawing uniformly within each category.
Factify-5WQA~\citep{DBLP:journals/corr/abs-2305-04329} has
sufficient Why-type entries—a property not shared by ZSRE or CounterFact.
\textbf{(3) Forget/Retain split.} Each label's 1,000 samples are split
equally: 500 forget, 500 retain.
Evaluation uses 100 samples per label
per set (500 forget $+$ 500 retain total), stratified to ensure equal
5W representation.
\textbf{(4) Format standardisation.} Each sample is formatted as a
\texttt{(question, answer, label, rephrases)} tuple compatible with
EasyEdit~\citep{DBLP:journals/corr/abs-2308-07269}.
The \texttt{label} reflects the \textit{semantic
type} of the relation queried, not the surface question word.

\paragraph{Sample Format.}
Each \fivewbench\ record is a JSON object containing the question, ground-truth
answer, the 5W label, and up to three rephrased question variants generated to
test robustness of editing methods to surface-form variation.
Table~\ref{tab:sample_example} shows a representative \texttt{what}-type
instance from the forget split, drawn from the ZSRE source.

\subsection{Why-Type Facts Are Structurally Different}

% \begin{table}[t]
% \centering
% \small
% \setlength{\tabcolsep}{6pt}
% \renewcommand{\arraystretch}{1.1}
% \caption{Per-category statistics in \fivewbench\ (1,000 samples per category split equally into 500 forget and 500 retain; 5,000 total samples). \texttt{Why}-type entries (amber) exhibit vastly longer answer spans and highly complex relational chains—causing severe gradient dilution during standard unlearning.}
% \label{tab:5wqa_stats}
% \begin{tabular}{lrrc}
% \toprule
% \textbf{Cat.} & \textbf{Ans.\ Len.} & \textbf{Multi-hop} & \textbf{Structure} \\
% \midrule
% Who   &  4.2 tok. &  1\% & Single-Entity \\
% What  &  4.8 tok. &  2\% & Single-Entity \\
% When  & 10.5 tok. &  2\% & Single-Entity \\
% Where &  4.2 tok. &  1\% & Single-Entity \\
% \rowcolor{warnamber}
% Why   & 40.1 tok. & 44\% & Causal Chain \\
% \midrule
% \textbf{Avg / Total} & \textbf{12.7 tok.} & \textbf{10\%} & \textbf{Mixed} \\
% \bottomrule
% \end{tabular}
% \end{table}

Why-type facts encode causal and relational chains (e.g., \textit{``Smoking
causes lung cancer because it introduces carcinogens into lung tissue''}).
Answer spans
average 40.1 tokens versus 4.2--10.5 for other types, and a staggering 44\% involve
multi-hop reasoning chains (Table~\ref{tab:5wqa_stats}, Appendix~\ref{app:5wqa_stats}).
This complexity means 
gradient ascent on a long token span cannot produce a coherent unlearning signal.
\fivewbench\ provides sufficient Why-type samples to study and quantify this failure 
mode for the first time.

% ══════════════════════════════════════════════════════════════════════════════
\section{The \maat\ Framework}
\label{sec:maat}

\begin{figure*}[t]
\centering
\includegraphics[width=\linewidth]{{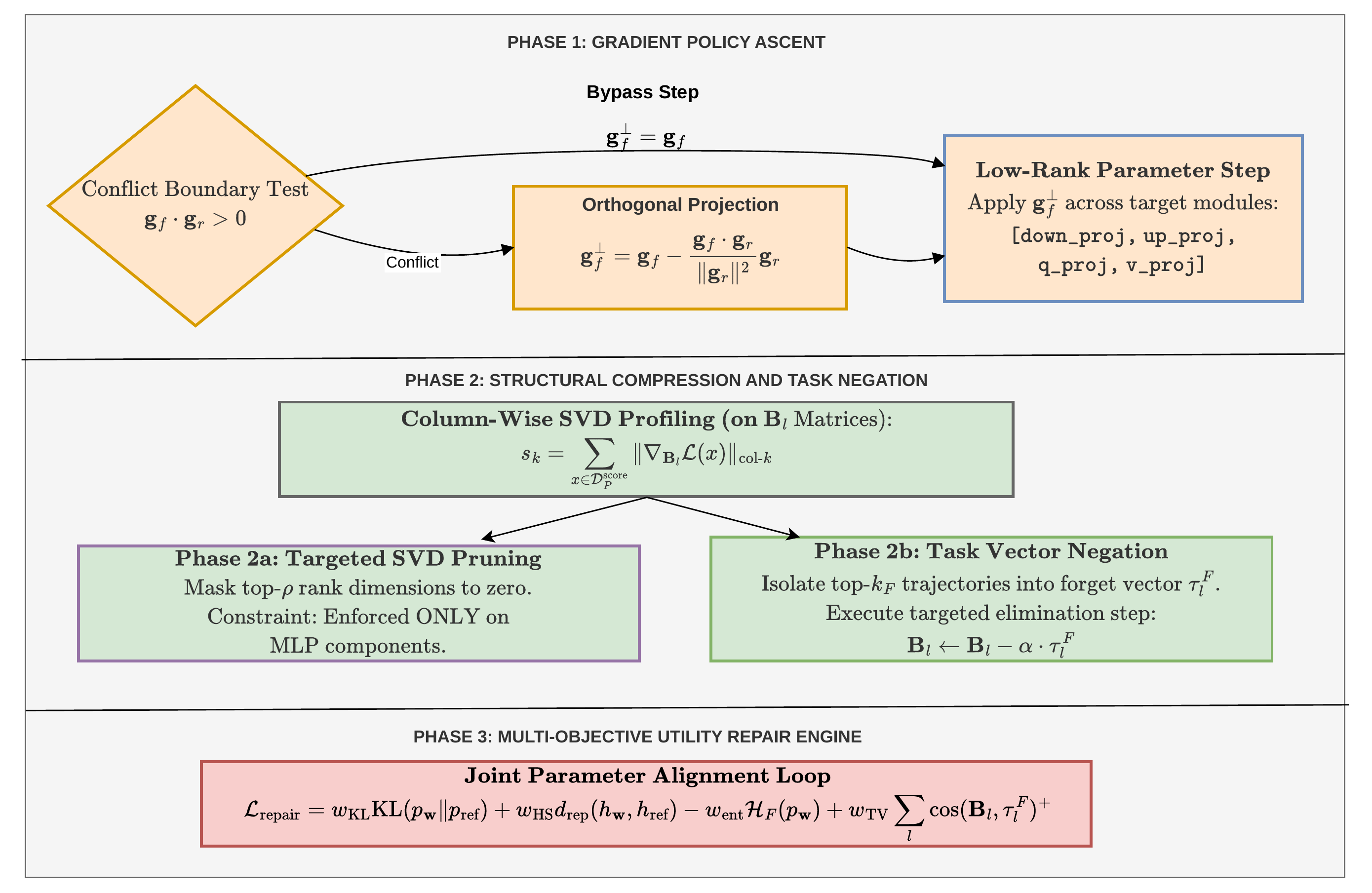}}
\caption{Overview of \maat\ ( multi-phase adapter-aware targeted unlearning ) architecture.}
\label{fig:maat_arch}
\end{figure*}

\maat\ addresses the unlearning challenge by operating on the
structure of the LoRA adapter's parameter space (Figure~\ref{fig:maat_arch}).
All three phases act exclusively on
adapter matrices $\{\mathbf{A}_l, \mathbf{B}_l\}$; base model weights
remain frozen throughout.

\subsection{Phase 1: Gradient-Projected Unlearning}
Standard gradient ascent applies a forget update $\mathbf{g}_f$ uniformly.
If $\mathbf{g}_f$ has components aligned with the retain gradient
$\mathbf{g}_r$, those components erode retained knowledge.
\maat\ removes
this component via \emph{conditional} orthogonal projection—applied only
when the forget and retain gradients actively conflict
($\mathbf{g}_f \cdot \mathbf{g}_r > 0$):
\begin{equation}
  \mathbf{g}_f^{\perp} =
  \begin{cases}
    \mathbf{g}_f
      - \dfrac{\mathbf{g}_f \cdot \mathbf{g}_r}{\|\mathbf{g}_r\|^2 + \epsilon}
      \,\mathbf{g}_r
      & \text{if } \mathbf{g}_f \cdot \mathbf{g}_r > 0 \\[6pt]
    \mathbf{g}_f & \text{otherwise}
  \end{cases}
  \label{eq:gradproject}
\end{equation}
where $\epsilon > 0$ is a small numerical stabiliser.
When the gradients do
not conflict, the full forget-ascent direction is preserved without
attenuation, maintaining unlearning signal strength.
The KL reference distribution is fixed to the pre-unlearning adapter state
$\mathbf{W}_{\text{ref}}$.
The unlearning objective to be minimised is:
\begin{equation}
  \mathcal{L}_{\text{unlearn}} =
    -\mathcal{L}_{\text{forget}}(q, a)
    \;+\;
\lambda\,\mathrm{KL}\!\left(
      p_{\mathbf{W}} \;\|\; p_{\mathbf{W}_{\text{ref}}}
    \right)_{\mathcal{D}_R}
  \label{eq:unlearn_obj}
\end{equation}
where $\lambda > 0$ controls the retain-anchoring strength.

\subsection{Phase 2a: SVD Rank-Dimension Pruning (MLP-only)}
\label{subsec: phase2a}

Residual forget-fact signal may persist in low-magnitude adapter directions
not captured by the projected gradient.
For each MLP LoRA pair
$(\mathbf{A}_l, \mathbf{B}_l)$, rank dimensions are scored by their
gradient-column norm over a sample of forget-set inputs:
\begin{equation}
  s_k = \sum_{x \in \mathcal{D}_F^{\text{score}}}
        \left\|\nabla_{\mathbf{B}_{l}}\mathcal{L}(x)\right\|_{\text{col-}k}
  \label{eq:svd_score}
\end{equation}

The top-$\rho$ fraction of rank dimensions by score are zeroed in both A
and B matrices on MLP modules (\texttt{down\_proj, up\_proj, gate\_proj})
only.
\textbf{Attention modules are excluded}: pruning attention rank
dimensions at non-trivial ratios destroys the instruction-following pathway.

\subsection{Phase 2b: Task Vector Negation}

Phase 2b introduces weight-space negation targeting the forget subspace.
The top-$k_F$ fraction of forget-scored rank dimensions of each LoRA B
matrix are masked to form a forget task vector $\bm{\tau}_l^F$, then subtracted:
\begin{equation}
  \mathbf{B}_l \gets \mathbf{B}_l - \alpha \cdot \bm{\tau}_l^F,
  \quad \alpha > 0
  \label{eq:taskvec_negate}
\end{equation}

By confining negation to the highest forget-scoring rank dimensions (here
$k_F = 50\%$), Phase 2b achieves targeted suppression without
inadvertently erasing retain-associated directions.

\subsection{Phase 3: Retain Repair with Hybrid Objective}

SVD pruning and gradient ascent may partially degrade retain performance.
Phase 3 recovers it through a hybrid repair
objective combining four terms:
\begin{multline}
  \mathcal{L}_{\text{repair}} = 
    w_{\text{KL}}\,\mathrm{KL}(p_{\mathbf{W}} \| p_{\text{ref}})_{\mathcal{D}_R} \\
    + w_{\text{HS}}\,d_{\text{rep}}(h_{\mathbf{W}}, h_{\text{ref}})_{\mathcal{D}_R} 
    - w_{\text{ent}}\,\mathcal{H}_F(p_{\mathbf{W}}) \\
    + w_{\text{TV}}\!\sum_l \cos(\mathbf{B}_l, \bm{\tau}_l^F)^+
  \label{eq:repair}
\end{multline}

where $d_{\text{rep}}(h, h') = 1 - \cos(h, h')$ is the hidden-state
representation distance, $\mathcal{H}_F(p_{\mathbf{W}})$ is the output
entropy of the current model evaluated on forget-set answer tokens,
and the final term penalises cosine similarity between current LoRA $\mathbf{B}_l$
weights and the forget task vector.
Crucially, the entropy term enters with a \emph{negative} sign:
minimising $-w_{\text{ent}}\mathcal{H}_F$ is equivalent to maximising
entropy on forget-set predictions, discouraging the model from recovering
forgotten content during the repair phase.

\subsection{Knowledge Implantation Protocol}

To benchmark unlearning in a controlled setting, we first implant target
facts via LoRA fine-tuning (rank $r$, scaling factor $\alpha_{\text{LoRA}}$,
targeting \texttt{q, k, v, o, gate, up, down} projections on layers
$[l_s, l_e]$) with 4-bit NF4 quantisation (learning rate $\eta_0 > 0$
for $E$ epochs).
Crucially, \textbf{the
LoRA adapter is not merged into the base model}: all \maat\ phases operate
directly on adapter weights.

\subsection{\maat{}  Algorithm}
We summarize the full \maat{} training procedure in Algorithm~\ref{alg:maat}.

\begin{algorithm}[t]
\caption{\maat: Three-Phase LoRA Adapter Unlearning}
\label{alg:maat}
\small
\begin{algorithmic}[1]
\Require Forget set $\mathcal{D}_F$; retain set $\mathcal{D}_R$;
         fine-tuned adapter $\{\mathbf{A}_l, \mathbf{B}_l\}$;
         frozen base $\mathbf{W}_{\text{base}}$;
 hyperparameters $T$, $S$, $\rho$, $\alpha$, $\eta_1$, $\eta_3$, $\eta_f$
\Ensure Unlearned adapter $\{\mathbf{A}_l', \mathbf{B}_l'\}$

\State \textbf{Save reference:} $\mathbf{W}_{\text{ref}} \gets \{\mathbf{A}_l, \mathbf{B}_l\}$
\State \textbf{-- Phase 1: Gradient-Projected Ascent --}
\For{$(q, a) \in \mathcal{D}_F$}
  \For{$t = 1$ to $T$}
    \State $\mathbf{g}_f \gets +\nabla\mathcal{L}(q,a)$;\quad
           $\mathbf{g}_r \gets \nabla\mathrm{KL}(p_\mathbf{W} \| p_{\mathbf{W}_\text{ref}})$
    \State Conditionally project $\mathbf{g}_f^\perp$ via Eq.~\ref{eq:gradproject};
 apply ascent step with $\eta_1$
  \EndFor
\EndFor
\State \textbf{-- Phase 2a: SVD Pruning (MLP LoRA only) --}
\State Score rank dims via Eq.~\ref{eq:svd_score};
 zero top-$\rho$
       in \texttt{down/up/gate\_proj} on layers $[l_s, l_e]$
\State \textbf{-- Phase 2b: Task Vector Negation --}
\State Score top-$k_F$ forget dims;
 compute $\bm{\tau}_l^F$;
       subtract $\alpha \cdot \bm{\tau}_l^F$ from $\mathbf{B}_l$
\State \textbf{-- Phase 3: Hybrid Repair --}
\For{$s = 1$ to $S$}
  \State Minimise $\mathcal{L}_{\text{repair}}$ (Eq.~\ref{eq:repair})
         across all 7 module types;
 cosine-decay $\eta_3 \to \eta_f$
\EndFor
\State \Return $\{\mathbf{A}_l', \mathbf{B}_l'\}$
\end{algorithmic}
\end{algorithm}
% ══════════════════════════════════════════════════════════════════════════════
\section{Experimental Setup}
\label{sec:setup}

\paragraph{Models.}
We evaluate on two instruction-tuned models from distinct architecture
families: \textbf{LLaMA 3.2-3B-Instruct}~\citep{DBLP:journals/corr/abs-2407-21783}
and \textbf{Gemma 3-4B-Instruct}~\citep{DBLP:journals/corr/abs-2403-08295}.
Both fit on a single consumer GPU ($\leq$24\,GB
VRAM). Inference: greedy decoding, temperature $= 0$,
\texttt{max\_new\_tokens} $= 100$. For hyperparameters on \fivewbench{} using \maat{}  refer to (Appendix~\ref{app:hparams}, Table~\ref{tab:hparams}).

\paragraph{Datasets.}
\textit{\fivewbench} (see \S\ref{sec:benchmark} for details): 1,000 per label (5,000 total); experiments use 100 per label per set (forget500 $+$ retain500),
stratified to ensure equal 5W representation.
TOFU~\citep{DBLP:journals/corr/abs-2401-06121}: We use the \textbf{forget05/retain95} split.
Although TOFU  is not statistically useful for label-wise semantic computation due to severe category imbalance (see Appendix~\ref{app:tofu_dist} for distributions), it is included to ensure comparability and completeness with existing literature. TOFU contains no meaningful Why-type questions.

\paragraph{Baselines.}
All baselines operate on the fine-tuned LoRA adapter.
\begin{itemize}[leftmargin=1.5em, itemsep=2pt]
  \item \textbf{Gradient Ascent (GA)}: Direct loss maximisation on the forget set.
  \item \textbf{Gradient Ascent with KL Divergence  (\klga{})}: GA regularised by KL divergence.
  \item \textbf{Adapter Negation (AN)}: Full task vector negation,
    $\alpha = 1.0$.
  \item \textbf{Retain-Only Fine-Tuning (RO-FT)}: Fine-tuning exclusively
    on the retain set.
\end{itemize}

\paragraph{Evaluation Metrics.}

\textit{LLM-as-Judge (primary).\citep{DBLP:journals/corr/abs-2306-05685}}
We evaluate Forget Success Rate (\fsr{}) and Retain Success Rate (\rsr{}) using a Qwen~2.5-7B judge that determines
whether the model's output semantically contains the ground-truth answer.
$\text{\fsr{}} = 1$ when the ground truth is \emph{not} present (successful
forgetting); $\text{\rsr{}} = 1$ when it is present (successful retention).
Full prompt template and evaluation rules are provided in
Appendix~\ref{app:judge_prompt}.

\textit{ROUGE}:
We report ROUGE-1, ROUGE-2, and ROUGE-L ~\citep{lin-2004-rouge}
on the forget and retain splits
before and after unlearning in the Appendix \ref{sec:ablation}.
Lower forget ROUGE indicates more successful token-level erasure; higher
retain ROUGE indicates better preservation of non-targeted knowledge.

\section{Results}
\label{sec:results}

Tables~\ref{tab:results_5wqa} and~\ref{tab:results_tofu} present results
on \fivewbench\ and TOFU respectively. Appendix~\ref{app:qualitative} 
provides qualitative examples illustrating how each method handles Why-type questions on both forget and retain splits.
\fivewbench's balanced 5W coverage exposes performance differences between
methods that TOFU's category-skewed evaluation cannot resolve.

\begin{table*}[t]
\centering
\footnotesize
\setlength{\tabcolsep}{3.5pt}
\caption{Unlearning results on \textbf{\fivewbench}\ (Factify; 100 forget $+$ 100 retain
per 5W label; 500 total each split). Judge: Qwen~2.5-7B.
\textbf{\fsr{}\,$\uparrow$}
= Forget Success Rate; \textbf{\rsr{}\,$\uparrow$} = Retain Success Rate.
\colorbox{warnamber}{Amber}: Why-type category row. \colorbox{wingreen}{Green}:
Best \fsr{}-\rsr{} balance per model.
Bold: \maat\ (proposed method).}
\label{tab:results_5wqa}
\begin{tabular}{ll rrrrrr rrrrrr}
\toprule
 & &
  \multicolumn{6}{c}{\textbf{\fsr{}{}{}{} (\%) $\uparrow$}} &
  \multicolumn{6}{c}{\textbf{\rsr{} (\%) $\uparrow$}} \\
\cmidrule(lr){3-8}\cmidrule(lr){9-14}
\textbf{Method} & \textbf{Model} &
  Who & What & When & Where & \cellcolor{warnamber}Why & Avg &
  Who & What & When & Where & \cellcolor{warnamber}Why & Avg \\
\midrule
\multicolumn{14}{l}{\textit{LLaMA 3.2-3B}} \\
\midrule
GA       & LLaMA &
  71.0 & 48.0 & 60.0 & 64.0 & \cellcolor{warnamber}44.0 & 57.4 &
  41.0 & 59.0 & 43.0 & 45.0 & \cellcolor{warnamber}78.0 & 53.2 \\
\klga{}    & LLaMA &
  43.0 & 27.0 & 38.0 & 39.0 & \cellcolor{warnamber}22.0 
 & 33.8 &
  65.0 & 74.0 & 53.0 & 64.0 & \cellcolor{warnamber}93.0 & 69.8 \\
AN       & LLaMA &
  99.0 & 100.0 & 100.0 & 100.0 & \cellcolor{warnamber}100.0 & 99.8 &
   0.0 &   1.0 &   0.0 &   1.0 & \cellcolor{warnamber}  0.0 &  0.4 \\
RO-FT    & LLaMA &
  80.0 & 74.0 & 82.0 & 79.0 & \cellcolor{warnamber}72.0 & 77.4 &
  36.0 & 40.0 & 23.0 & 35.0 & \cellcolor{warnamber}42.0 & 35.2 \\
\rowcolor{wingreen}
\textbf{\maat} & \textbf{LLaMA} &
  \textbf{83.0} & \textbf{79.0} 
 & \textbf{82.0} & \textbf{80.0} &
  \textbf{\cellcolor{wingreen}63.0} & \textbf{77.4} &
  \textbf{72.0} & \textbf{77.0} & \textbf{71.0} & \textbf{73.0} &
  \textbf{\cellcolor{wingreen}65.0} & \textbf{71.6} \\
\midrule
\multicolumn{14}{l}{\textit{Gemma 3-4B}} \\
\midrule
GA       & Gemma &
  82.0 & 58.0 & 73.0 & 50.0 & \cellcolor{warnamber}43.0 & 61.2 &
  51.0 & 59.0 & 24.0 & 51.0 & \cellcolor{warnamber}70.0 & 51.0 \\
\klga{}    & Gemma &
  60.0 & 35.0 & 53.0 & 30.0 & \cellcolor{warnamber}36.0 & 42.8 &
  62.0 & 70.0 & 39.0 & 66.0 & \cellcolor{warnamber}71.0 & 61.6 \\
AN       & 
 Gemma &
  92.0 & 83.0 & 82.0 & 78.0 & \cellcolor{warnamber}75.0 & 82.0 &
  19.0 & 32.0 & 19.0 & 21.0 & \cellcolor{warnamber}34.0 & 25.0 \\
RO-FT    & Gemma &
  63.0 & 45.0 & 58.0 & 43.0 & \cellcolor{warnamber}48.0 & 51.4 &
  82.0 & 79.0 & 47.0 & 81.0 & \cellcolor{warnamber}63.0 & 70.4 \\
\rowcolor{wingreen}
\textbf{\maat} & \textbf{Gemma} &
  \textbf{77.0} & \textbf{61.0} & \textbf{68.0} & \textbf{59.0} &
  \textbf{\cellcolor{wingreen}55.0} & \textbf{64.0} &
  \textbf{62.0} & \textbf{69.0} & \textbf{52.0} & \textbf{70.0} &
  \textbf{\cellcolor{wingreen}56.0} & \textbf{61.8} \\
\bottomrule
\end{tabular}
\end{table*}

% \subsection{\fivewbench\ Results: \fsr-\rsr{}{} Tradeoff Analysis}

% When evaluating label-wise unlearning efficiency (typically considered as the harmonic mean balance between \fsr{} and \\rsr{}{}), \maat\ achieves the best \textbf{aggregate \fsr-\rsr{}{} balance} on both models. 
% On LLaMA, \maat\ reaches 77.4\% \fsr{} and 71.6\% \rsr{}—$+20\%$
% \fsr{} over GA (57.4\%) while also improving \rsr{} by $+18.4\%$ (53.2\%
% $\to$ 71.6\%).
% On Gemma, \maat\ achieves 64.0\% \fsr{} and 61.8\% \\rsr{}{}.

% \textbf{Baseline tradeoff failure modes.} GA achieves moderate \fsr{}
% (57.4\% LLaMA, 61.2\% Gemma) at the cost of poorer \rsr{} than \maat.
% \klga{} achieves higher \rsr{} (69.8\% LLaMA) but substantially reduces \fsr{}
% to 33.8\%---sacrificing unlearning effectiveness for retention.
% Adapter Negation (AN) achieves high \fsr{} (99.8\% LLaMA, 82.0\% Gemma)
% but \rsr{} collapses catastrophically to 0.4\% and 25.0\% respectively.

% \textbf{Why-type specifics.} On the Why category, \maat\ achieves 63\%
% \fsr{} / 65\% \rsr{} on LLaMA and 55\% \fsr{} / 56\% \rsr{} on Gemma.
% Compared
% to GA (44\% \fsr{} / 78\% \rsr{} on LLaMA), \maat\ gains $+19\%$ \fsr{} while
% retaining comparable \rsr{}.
% Compared to \klga{} (22\% \fsr{} / 93\% \rsr{}),
% \maat\ more than doubles \fsr{} at an acceptable \rsr{} cost.
% An \fsr{} above
% 60\% with \rsr{} above 60\% on the most difficult category—Why—represents
% a qualitatively new operating point that no baseline achieves.

\begin{table}[t]
\centering
\small
\setlength{\tabcolsep}{5pt}
\caption{Unlearning results on \textbf{TOFU}
(forget05/retain95: 200 forget, 3{,}800 retain).
Judge: Qwen~2.5-7B. TOFU contains no Why-type questions (N/A).
\colorbox{wingreen}{Green}: \maat\ results.}
\label{tab:results_tofu}
\begin{tabular}{llcc}
\toprule
\textbf{Method} & \textbf{Model} & \textbf{\fsr{} (\%) $\uparrow$}
  & \textbf{\rsr{} (\%) $\uparrow$} \\
\midrule
\multicolumn{4}{l}{\textit{LLaMA 3.2-3B}} \\
\midrule
GA       & LLaMA & 68.0 & 32.0 \\
\klga{}    & LLaMA & 57.5 & 39.6 \\
AN       & LLaMA & \textbf{100.0} & 0.2 \\
RO-FT    & LLaMA & 67.0 & 45.1 \\
\rowcolor{wingreen}
\textbf{\maat} & \textbf{LLaMA} & \textbf{67.5} & \textbf{46.6} \\
\midrule
\multicolumn{4}{l}{\textit{Gemma 3-4B}} \\
\midrule
GA       & Gemma & 53.0 & 49.7 \\
\klga{}    & Gemma & 54.0 & 47.4 \\
AN      
       & Gemma & 62.0 & 33.8 \\
RO-FT    & Gemma & 63.0 & 46.2 \\
\rowcolor{wingreen}
\textbf{\maat} & \textbf{Gemma} & \textbf{61.5} & \textbf{48.7} \\
\bottomrule
\end{tabular}
\end{table}

% \subsection{TOFU Results (forget05/retain95)}

% On TOFU, \maat\ achieves the best \rsr{} among methods with $> 60\%$ \fsr{}
% on both models.
% On LLaMA: AN reaches 100\% \fsr{} but completely destroys
% retention (\rsr{} $= 0.2\%$), while \maat\ achieves 67.5\% \fsr{} with 46.60\%
% \rsr{}—the highest retention rate among competitive forgetting methods.
% \klga{} on LLaMA shows lower \fsr{} (57.5\%) with similar \rsr{} costs.
% On Gemma, methods cluster more closely.
% \maat's \rsr{} of 48.7\% exceeds \klga{}'s
% 47.4\% and RO-FT's 46.2\%, demonstrating a consistent retention advantage
% at comparable \fsr{}.

\subsection{\fivewbench\  Results (forget500/retain500)}

\paragraph{MAAT dominates the forget--retain Pareto frontier.}
\maat\ achieves the best aggregate \fsr--\rsr{} balance on both models:
$77.4\%$ / $71.6\%$ on \llamamodel\ and $64.0\%$ / $61.8\%$ on
\gemmamodel.
The cleanest single comparison is against RO-FT: both methods reach
identical $77.4\%$ average \fsr{} on \llamamodel, yet \maat\ delivers
this with $71.6\%$ \rsr{} versus RO-FT's $35.2\%$---a $+36.4$-point
retention gain at zero forget cost.
More strikingly, \maat\ is the \textit{only} method to simultaneously
exceed $60\%$ \fsr{} and $60\%$ \rsr{} on all five 5W categories on
\llamamodel\ (Who: $83/72$, When: $79/77$, What: $82/71$, Where:
$80/73$, Why: $63/65$); no baseline achieves this threshold on any
single category.

\paragraph{Baseline tradeoff failure modes.}
Each baseline occupies a distinct failure region of the Pareto frontier.
GA achieves moderate average \fsr{} ($57.4\%$ \llamamodel, $61.2\%$
\gemmamodel) at the cost of \rsr{} lagging \maat\ by $18$ and $11$
points respectively.
GA+KL ($33.8\%$ \fsr{} on \llamamodel) sacrifices forgetting for
retention---the lowest average \fsr{} of any real method.
Adapter Negation (AN) maximises \fsr{} ($99.8\%$ \llamamodel, $82.0\%$
\gemmamodel) but triggers catastrophic forgetting uniformly across
\textit{all} 5W categories: \rsr{} collapses to $0.4\%$ on \llamamodel\
and $25.0\%$ on \gemmamodel.
This confirms that full task vector negation removes both forget and
retain knowledge indiscriminately---the entire adapter delta encodes
both simultaneously.
The partial resistance on \gemmamodel\ (25.0\% \rsr{}) compared to \llamamodel\ (0.4\% \rsr{}) suggests architecture-dependent sensitivity to full adapter negation.

\paragraph{Why-type results and the under-forgetting artifact.}
On the Why category, \maat\ achieves $63\%$ \fsr{} / $65\%$ \rsr{} on
\llamamodel\ and $55\%$ \fsr{} / $56\%$ \rsr{} on \gemmamodel.
GA's higher Why-\rsr{} ($78\%$ on \llamamodel) is not evidence of
better retention---it is a mechanical artifact of under-forgetting.
At $44\%$ Why-\fsr, more than half the target causal facts remain in
the model, so retention is preserved by default rather than by design.
\maat\ forgets $19$ percentage points more Why-type knowledge ($63\%$
vs $44\%$) while losing only $13$ \rsr{} points ($65\%$ vs $78\%$)---
the only method that meaningfully removes causal knowledge without
indiscriminate retain damage.
GA+KL ($22\%$ Why-\fsr{} / $93\%$ Why-\rsr{} on \llamamodel)
represents the extreme of this artifact: near-perfect retention because
almost nothing has been forgotten.

\paragraph{Per-category localized failures.}
Beyond the Why category, \fivewbench's balanced coverage reveals
localized failures in other categories.
GA's What-type \fsr{} on \llamamodel\ ($48\%$) approaches its Why-type
\fsr{} ($44\%$), and \maat's \fsr{} gap over GA is actually larger on
What ($+31$ points) than on Why ($+19$ points)---showing that GA's
weakness is not unique to causal knowledge.
GA+KL's When-\rsr{} on \gemmamodel\ collapses to $39\%$ while all
other categories remain $62$--$71\%$, indicating a localized temporal
knowledge failure under KL regularization on this architecture.
RO-FT's Where-\rsr{} on \llamamodel\ ($23\%$) is dramatically lower
than its other categories ($35$--$42\%$).
\fivewbench's balanced $500$-sample splits per category provide the statistical power to detect these localized failures---a granularity unavailable in existing benchmarks.

\paragraph{Consistency and architecture sensitivity.}
\maat's \fsr{} ranges $63$--$83\%$ across categories on \llamamodel,
compared to GA's $44$--$71\%$, indicating more consistent unlearning
behavior regardless of question type---an important practical property
since real-world forget requests may contain any mix of knowledge types.
All methods degrade moving from \llamamodel\ to \gemmamodel, but
differently: RO-FT swings dramatically (\fsr{} $-26.0$, \rsr{}
$+35.2$), indicating high architecture-dependence, while \maat\
degrades more uniformly (\fsr{} $-13.4$, \rsr{} $-9.8$), maintaining
a balanced operating point across both architectures.

\subsection{TOFU Results (forget05/retain95)}

On TOFU, \maat\ achieves the highest \rsr{} among methods with
$>60\%$ \fsr{} on \llamamodel\ ($67\%$ \fsr{} / $46.6\%$ \rsr).
AN again triggers catastrophic forgetting ($100\%$ \fsr{} / $0.2\%$
\rsr), replicating the indiscriminate-negation pattern seen on
\fivewbench.
On \gemmamodel, methods cluster tightly: \maat\ ($61.5\%$/$48.7\%$),
RO-FT ($63.0\%$/$46.2\%$), and GA+KL ($54.0\%$/$47.4\%$) differ by
$1$--$3$ points---within the noise range of LLM-judge evaluation.
TOFU's category imbalance ($84.7\%$ What-type, $\leq 5\%$ each for
When/Who/Where, $1.2\%$ Why) means aggregate \fsr--\rsr{} is dominated
by a single question type.
\fivewbench's balanced $500$-sample splits per category provide the
statistical power to surface per-category differences that TOFU's
distribution cannot resolve.
TOFU results are included for comparability with existing literature.

% ══════════════════════════════════════════════════════════════════════════════
\section{Conclusion}

We introduced \fivewbench, a balanced 5,000-sample benchmark that makes
causal unlearning failures quantifiable for the first time, and \maat, a
three-phase structured LoRA adapter unlearning framework that concentrates
the forgetting signal on rank dimensions specifically activated by
forget-set inputs.

Using \fivewbench, we established that no existing baseline simultaneously
achieves high forgetting and high retention on Why-type causal
knowledge---and that \maat\ is the first method to do so, exceeding $60\%$
\fsr{} and $60\%$ \rsr{} on all five 5W categories on \llamamodel,
including Why-type causal questions where all baselines face a fundamental
forget--retain tradeoff.

The performance gap between \llamamodel\ ($77.4/71.6$) and \gemmamodel\
($64.0/61.8$) reveals that adapter-based unlearning inherits the base
model's knowledge encoding structure: forget and retain knowledge are more
separable in \llamamodel's adapter rank dimensions, while \gemmamodel's
interleaved local/global attention pattern distributes knowledge more
diffusely, reducing separability. Architecture-aware adapter rank selection
is an important open direction for future work.

Beyond unlearning, \fivewbench's balanced 5W taxonomy and
EasyEdit-compatible format make it directly applicable to model editing
evaluation---including insertion and modification on causal knowledge, a
category no existing editing benchmark adequately covers.
% ── Limitations ───────────────────────────────────────────────────────────────
\section*{Limitations}

\paragraph{Evaluation protocol.}
\fsr{} and \rsr{} are computed using a single Qwen~2.5-7B LLM judge.
While this provides a reproducible, semantically-aware alternative to
exact-match metrics, it may have calibration differences compared to
proprietary judges or human evaluation. Judge availability and versioning
may affect reproducibility across time. Future work should complement
LLM-judge evaluation with adversarial paraphrase probing to assess whether
unlearning is genuine or merely suppresses verbatim recall.

\paragraph{Dataset scope.}
\fivewbench\ is derived from the Factify-5WQA corpus~\citep{DBLP:journals/corr/abs-2305-04329}, which was
designed for fact verification rather than knowledge editing. The
subject-extraction pipeline may introduce noise for facts with implicit
or pronominal subjects. Answer length and multi-hop statistics
in Table~\ref{tab:5wqa_stats} are computed on the full annotated dataset. The complete benchmark will be released upon publication.

\paragraph{Model and benchmark scope.}
Evaluation covers two model families at 3--4B scale on \fivewbench\ and
TOFU; extension to 7B+ checkpoints and additional architectures is left
for future work. The performance gap between \llamamodel\ and \gemmamodel\ suggests that adapter rank separability is architecture-dependent, and optimal hyperparameters for Phase~2a pruning ratio $\rho$ and Phase~2b mask $k_F$ may require architecture-specific tuning. While \fivewbench's format supports insertion and modification operations, this paper evaluates only unlearning; extending to other editing operations is left for future work.
% ── Ethics Statement ──────────────────────────────────────────────────────────
\section*{Ethical considerations}
The unlearning methods developed in this paper address legitimate 
privacy and safety needs including GDPR compliance and correction 
of harmful associations. We acknowledge a dual-use concern: the 
same techniques could be applied adversarially to remove 
safety-relevant knowledge from aligned models.

\bibliography{references}

\appendix

\section{Ablation Study}
\label{sec:ablation}

\begin{figure*}[t]
\centering
\includegraphics[width=\textwidth]{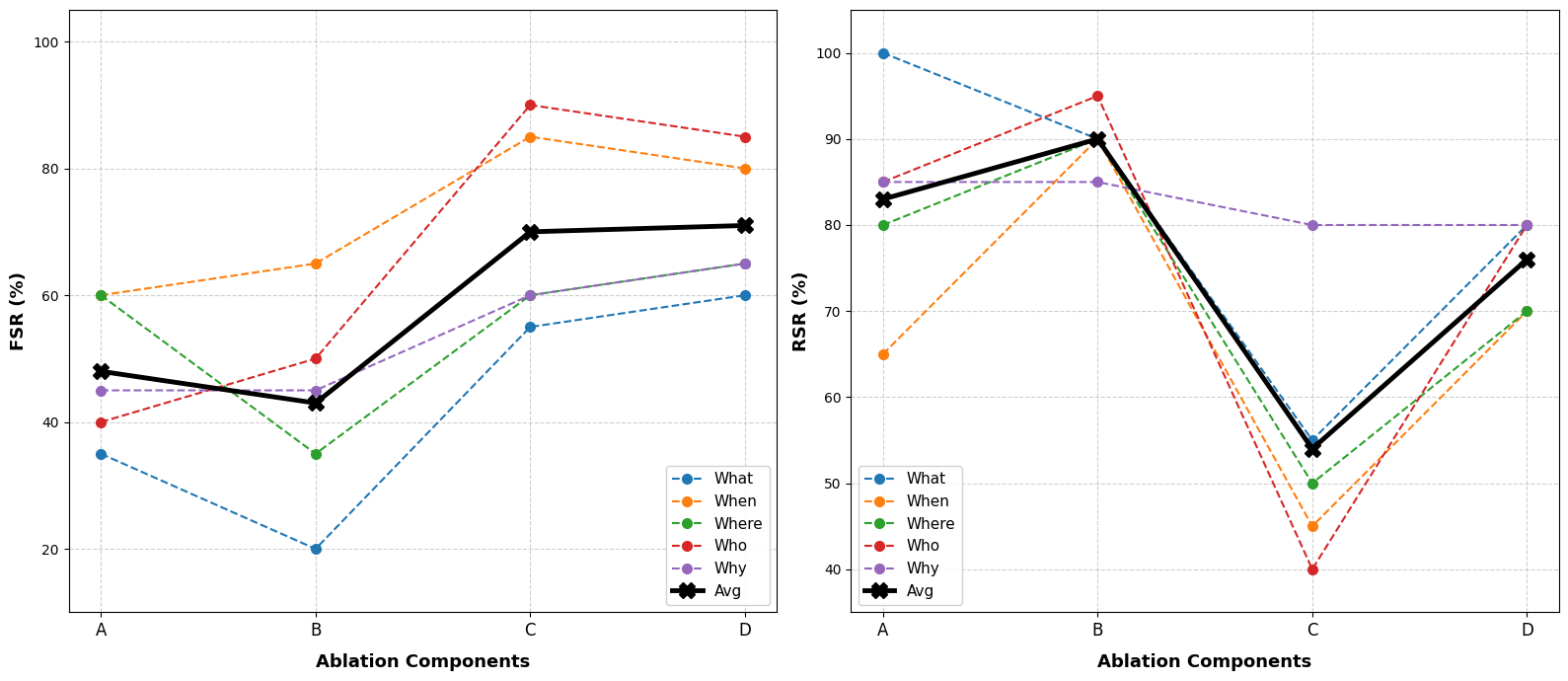}
\caption{Ablation study on \fivewbench\ (\llamamodel; 200 samples:
20 forget $+$ 20 retain per label). \fsr{} $\uparrow$, \rsr{} $\uparrow$.
\textbf{A}: Gradient projection $+$ SVD (MLP only) $+$ KL-only repair.
\textbf{B}: A with hybrid repair (all four loss terms).
\textbf{C}: B $+$ attention pruning ($\rho_\text{attn} = 0.01$).
\textbf{D}: Full \maat\ (B $+$ task vector negation; no attention pruning).}
\label{fig:ablation}
\end{figure*}

To understand the contribution of each \maat\ phase, we conduct ablation
experiments on a 200-sample \fivewbench\ subset (20 forget $+$ 20 retain
per label) on \llamamodel. We first confirm successful knowledge implantation
(Table~\ref{tab:finetune_rouge}), then evaluate four ablation conditions
(Figure~\ref{fig:ablation}) that progressively
introduce \maat\ components. Post-unlearning ROUGE scores
(Table~\ref{tab:post_rouge}) complement the FSR/RSR analysis by quantifying
token-level erasure and preservation.

% ---- Pre-unlearning ROUGE (knowledge implantation check) -------------------

\paragraph{Knowledge implantation verification.}
Table~\ref{tab:finetune_rouge} reports ROUGE scores of the fine-tuned LoRA
adapter \emph{before} any unlearning on both \fivewbench\ splits. High scores on
the forget set confirm that target facts have been successfully implanted;
high scores on the retain set confirm that general knowledge is intact.
These serve as upper-bound references: lower forget ROUGE after unlearning
indicates successful erasure, while retain ROUGE should remain as close to
these baselines as possible.

\begin{table}[h]
\centering
\small
\setlength{\tabcolsep}{4.5pt}
\caption{ROUGE scores of the fine-tuned LoRA adapter (pre-unlearning)
on \fivewbench\ forget and retain sets. High scores on both splits confirm
successful knowledge implantation prior to unlearning.}
\label{tab:finetune_rouge}
\begin{tabular}{llcccc}
\toprule
\textbf{Model} & \textbf{Set} & \textbf{R-1} & \textbf{R-2}
  & \textbf{R-L} & \textbf{R-Lsum} \\
\midrule
LLaMA 3.2-3B & Forget & 0.817 & 0.599 & 0.811 & 0.812 \\
LLaMA 3.2-3B & Retain & 0.979 & 0.696 & 0.980 & 0.979 \\
\midrule
Gemma 3-4B   & Forget & 0.786 & 0.509 & 0.776 & 0.776 \\
Gemma 3-4B   & Retain & 0.813 & 0.498 & 0.803 & 0.805 \\
\bottomrule
\end{tabular}
\end{table}

% ---- Ablation conditions ---------------------------------------------------

\paragraph{Ablation conditions.}
Each condition adds or modifies one component relative to the previous,
isolating its effect on the forget--retain tradeoff.

\textbf{Condition~A} (Gradient projection $+$ SVD $+$ KL-only repair;
no task vector negation, no attention pruning):
Phase~1 gradient-projected ascent and Phase~2a SVD rank-dimension pruning
on MLP modules only, paired with a repair phase that uses only the
KL-divergence term from Eq.~\ref{eq:repair}.
Achieves 48\% \fsr{} and 83\% \rsr{}.
The strong retention reflects effective anchor protection from gradient
projection; however, the absence of structural erasure components beyond
MLP pruning limits overall unlearning efficacy.
Notably, a \texttt{Why}-\fsr{} of 45\% indicates that gradient projection
alone provides a stable baseline for unlearning causal chains.

\textbf{Condition~B} (Condition~A $+$ hybrid repair;
no task vector negation, no attention pruning):
Same as Condition~A, but Phase~3 is upgraded from KL-only to the full hybrid
repair objective (Eq.~\ref{eq:repair}), incorporating the KL-divergence,
hidden-state representation distance, and negative forget-set entropy terms.
Drops \fsr{} to 43\% while driving \rsr{} up to 90\%.
The richer repair phase over-corrects: without a structural erasure anchor,
it dominates the parameter updates and inadvertently recovers forgotten
content.

\textbf{Condition~C} (Condition~B $+$ attention pruning;
no task vector negation):
Extends Condition~B by applying SVD rank-dimension pruning to attention
modules (q\_proj, k\_proj, v\_proj, o\_proj) at a pruning ratio of
$\rho_\text{attn} = 0.01$, in addition to MLP pruning.
Sharply shifts the balance: \fsr{} jumps to 70\% but \rsr{} drops to 54\%.
Structural attention pruning strongly suppresses forget-set knowledge,
but induces severe collateral damage to retained distributions---empirically
confirming the design decision in \S\ref{subsec: phase2a} to exclude attention
modules from pruning.

\textbf{Condition~D} (Full \maat):
Removes attention pruning from Condition~C and instead introduces Phase~2b
task vector negation on the top-$k_F$ forget-scored rank dimensions,
alongside the full hybrid repair phase (Phase~3) including the task-vector
cosine penalty term.
The complete pipeline: Phase~1 (gradient-projected ascent) $+$ Phase~2a
(SVD pruning, MLP only) $+$ Phase~2b (task vector negation) $+$ Phase~3
(full hybrid repair with all four loss terms).
Achieves 71\% \fsr{} and 76\% \rsr{}.
Task vector negation provides comparable forgetting strength to attention
pruning (71\% vs.\ 70\% \fsr{}) with far better retention (76\% vs.\ 54\%
\rsr{}), serving as the critical differentiator that reclaims retain
performance while maintaining effective unlearning---particularly on
multi-hop causal structures (\texttt{Why}-\fsr{} = 65\%).

% ---- Ablation figure -------------------------------------------------------

% ---- Ablation FSR/RSR table (optional — uncomment if desired) --------------

% \begin{table}[htpb]
% \centering
% \small
% \setlength{\tabcolsep}{4pt}
% \renewcommand{\arraystretch}{1.1}
% \caption{\maat\ ablation on \fivewbench\ (\llamamodel; 200 samples:
% 20 forget $+$ 20 retain per label).
% \fsr{} $\uparrow$, \rsr{} $\uparrow$.}
% \label{tab:ablation}
% \begin{tabular}{l cccc c cccc}
% \toprule
%  & \multicolumn{4}{c}{\textbf{\fsr{} (\%) $\uparrow$}} & & \multicolumn{4}{c}{\textbf{\rsr{} (\%) $\uparrow$}} \\
% \cmidrule{2-5} \cmidrule{7-10}
% \textbf{Cat.} & A & B & C & D & & A & B & C & D \\
% \midrule
% What  & 35 & 20 & 55 & 60 & & 100 & 90 & 55 & 80 \\
% When  & 60 & 65 & 85 & 80 & & 65  & 90 & 45 & 70 \\
% Where & 60 & 35 & 60 & 65 & & 80  & 90 & 50 & 70 \\
% Who   & 40 & 50 & 90 & 85 & & 85  & 95 & 40 & 80 \\
% \rowcolor{warnamber}
% Why   & 45 & 45 & 60 & 65 & & 85  & 85 & 80 & 80 \\
% \midrule
% \textbf{Avg} & \textbf{48} & \textbf{43} & \textbf{70} & \textbf{71} & & \textbf{83} & \textbf{90} & \textbf{54} & \textbf{76} \\
% \bottomrule
% \end{tabular}
% \end{table}

% ---- Post-unlearning ROUGE -------------------------------------------------

\paragraph{Post-unlearning ROUGE analysis.}
Table~\ref{tab:post_rouge} reports ROUGE scores on the forget and retain
splits \emph{after} applying each ablation condition, complementing the
FSR/RSR analysis with token-level overlap metrics. Lower forget ROUGE
indicates more successful knowledge erasure; higher retain ROUGE indicates
better preservation of non-targeted knowledge. Compared to the
pre-unlearning baselines in Table~\ref{tab:finetune_rouge}, these values
quantify the retain--forget ROUGE tradeoff induced by each component.

Condition~A (KL-only repair) provides a balanced initial baseline,
maintaining a high retain R-1 (0.859) due to effective anchor protection.
Upgrading to hybrid repair in Condition~B further elevates retain
preservation to an experimental maximum (R-1 = 0.892), but over-corrects:
forget R-1 inflates to 0.614, confirming that the repair phase recovers
forgotten content without a structural erasure anchor. Condition~C sharply
reverses this dynamic: attention pruning drives forget R-1 down to 0.325,
but incurs heavy collateral damage (retain R-1 drops to 0.644). Full
\maat\ (Condition~D) resolves this tradeoff: by pairing MLP pruning with
task vector negation instead of attention pruning, it maintains deep
forget-set erasure (R-1 = 0.319) while recovering retain performance
(R-1 = 0.721)---a $+$0.077 retain improvement over Condition~C at
equivalent forget-set erasure depth.

\begin{table}[htpb]
\centering
\small
\setlength{\tabcolsep}{4pt}
\renewcommand{\arraystretch}{1.1}
\caption{Post-unlearning ROUGE scores per ablation condition
(\llamamodel; 200 samples: 20 forget $+$ 20 retain per label).
\textbf{A}: Gradient projection $+$ SVD (MLP) $+$ KL-only repair.
\textbf{B}: A with hybrid repair.
\textbf{C}: B $+$ attention pruning ($\rho_\text{attn} = 0.01$).
\textbf{D}: Full \maat\ (B $+$ task vector negation; no attention pruning).}
\label{tab:post_rouge}
\begin{tabular}{llccc}
\toprule
\textbf{Condition} & \textbf{Split} & \textbf{R-1} & \textbf{R-2} & \textbf{R-L} \\
\midrule
\multirow{2}{*}{Cond.\,A (KL-only Repair)}
  & Forget & 0.520 & 0.376 & 0.511 \\
  & Retain & 0.859 & 0.529 & 0.854 \\
\midrule
\multirow{2}{*}{Cond.\,B (Hybrid Repair)}
  & Forget & 0.614 & 0.369 & 0.601 \\
  & Retain & 0.892 & 0.646 & 0.890 \\
\midrule
\multirow{2}{*}{Cond.\,C ($+$ Attn Pruning)}
  & Forget & 0.325 & 0.189 & 0.314 \\
  & Retain & 0.644 & 0.414 & 0.639 \\
\midrule
\multirow{2}{*}{Cond.\,D (Full \maat)}
  & Forget & 0.319 & 0.178 & 0.311 \\
  & Retain & 0.721 & 0.453 & 0.718 \\
\bottomrule
\end{tabular}
\end{table}

\section{Complete Unlearning Metric Profiles Across All Methods}
\label{app:metric_profiles}

\begin{figure*}[h]
\centering
\includegraphics[width=\textwidth]{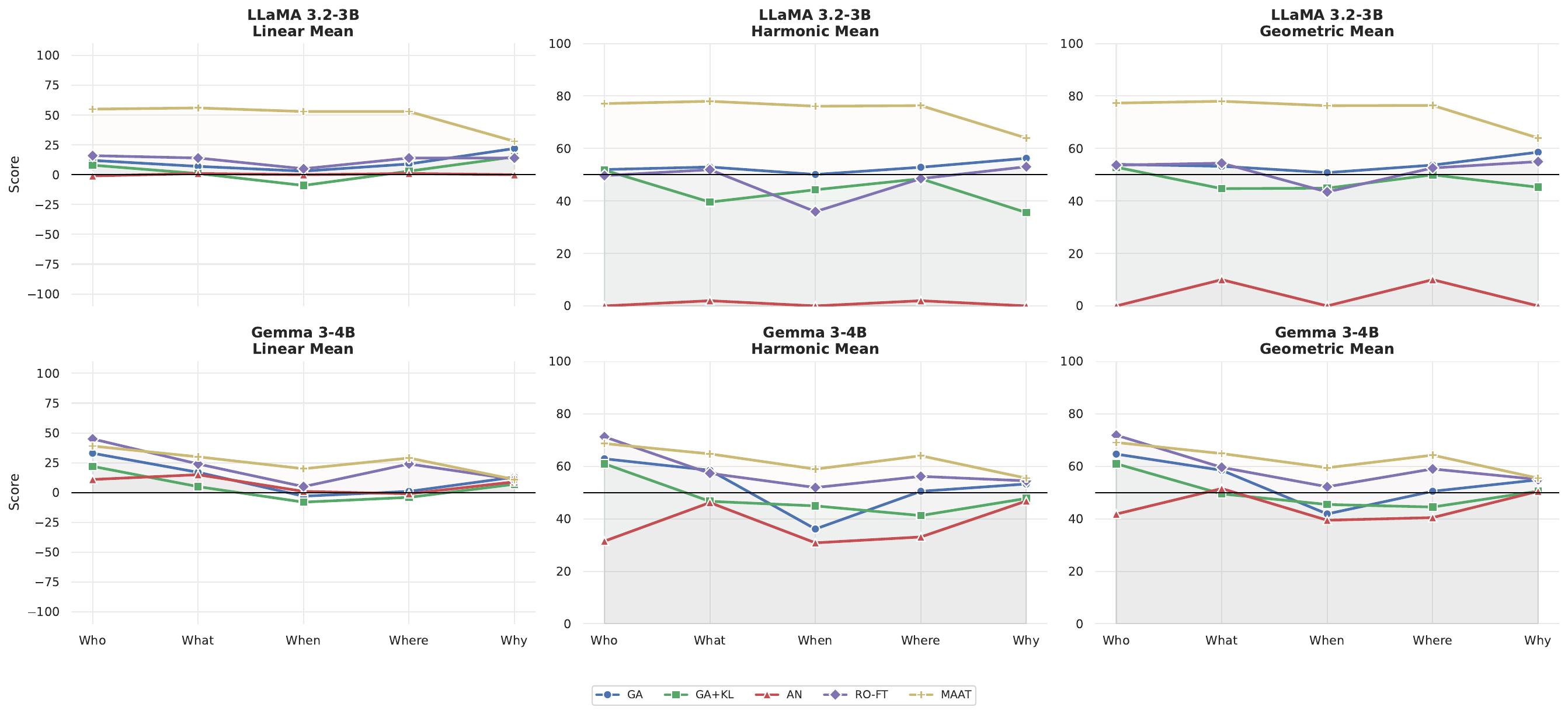}
\caption{Complete unlearning metric profiles per 5W label for LLaMA~3.2-3B
(top row) and Gemma~3-4B (bottom row), across three aggregation metrics:
Linear, Geometric, and Harmonic.
\maat\ (gold) achieves the highest or near-highest Geometric and Harmonic mean
across all categories on both models.
Adapter Negation (AN, red) reaches near-zero Geometric and Harmonic scores due
to catastrophic retention collapse, despite high forgetting.
\klga{} (green) maintains high Geometric/Harmonic scores on some categories
via strong retention preservation but under-forgets relative to \maat{}.
The \textit{When} category exhibits the lowest scores across most methods and
metrics, consistent with the short temporal-anchor answers that resist
gradient-ascent disruption.}
\label{fig:metric_profiles}
\end{figure*}

Figure~\ref{fig:metric_profiles} presents the full per-label unlearning metric
profiles for all five methods on both LLaMA~3.2-3B and Gemma~3-4B, evaluated
across three complementary aggregation functions:

% \begin{itemize}[leftmargin=1.5em, itemsep=2pt]
%   \item \textbf{Linear Mean:} Positive values indicate net forgetting gains exceed retention losses. This metric is symmetric and centred at zero, making it easy to interpret directional imbalances.
%   \item \textbf{Geometric Mean:} Penalises moderate imbalance and collapses to zero if either metric is zero. Methods with catastrophically low retention score near zero despite high forgetting.
%   \item \textbf{Harmonic Mean:} The most conservative aggregator that heavily penalises extreme imbalances between forgetting and retention. 
% \end{itemize}

\begin{itemize}
\item \textbf{Linear Mean}: ($\text{FSR} + \text{RSR} - 100$). 
Centred at zero; positive values indicate net forgetting gains exceed 
retention losses.

\item \textbf{Geometric Mean}: ($\sqrt{\text{FSR} \times \text{RSR}}$). 
Collapses to zero if either metric is zero, penalising catastrophic 
imbalance.

\item \textbf{Harmonic Mean}: ($\frac{2 \times \text{FSR} \times \text{RSR}}{\text{FSR} + \text{RSR}}$). 
The most conservative aggregator, heavily penalising extreme imbalances.
\end{itemize}

\noindent\textbf{Key observations from the full profiles.}
Across all three metrics, \maat\ consistently occupies the Pareto-dominant
region---achieving both high forgetting and high retention---while each
baseline exhibits a characteristic failure mode visible in the metric
profiles.
AN's Geometric and Harmonic scores collapse near zero on LLaMA due to near-zero retention, even though its Linear score is moderately positive.
This illustrates why single-metric or forget-only reporting conceals
retention failures.
\klga{}'s Geometric and Harmonic means are competitive on \textit{Who} and
\textit{Why} on LLaMA where its high retention partially compensates for low
forgetting, but its Linear scores are negative, revealing that retention gains
come at the price of under-forgetting.
GA's profile is the most consistent across metrics, but remains below \maat\
on all three aggregators.
The \textit{When} category is the hardest across both models and all methods,
corroborating the short-answer, single-entity structure that limits
gradient-ascent effectiveness.

\section{Per-Category Statistics}
\label{app:5wqa_stats}

Table~\ref{tab:5wqa_stats} summarizes the per-category characteristics of \fivewbench. Notably, \texttt{Why}-type questions exhibit substantially longer answers and higher multi-hop complexity, making them particularly challenging for standard unlearning methods.

\begin{table}[h]
\centering
\small
\setlength{\tabcolsep}{6pt}
\renewcommand{\arraystretch}{1.1}
\caption{Per-category statistics in \fivewbench\ (1,000 samples per category
split equally into 500 forget and 500 retain; 5,000 total samples).
\texttt{Why}-type entries (amber) exhibit vastly longer answer spans and
highly complex relational chains---causing severe gradient dilution during
standard unlearning.}
\label{tab:5wqa_stats}
\begin{tabular}{lrrc}
\toprule
\textbf{Cat.} & \textbf{Ans.\ Len.} & \textbf{Multi-hop} & \textbf{Structure} \\
\midrule
Who   &  4.2 tok. &  1\% & Single-Entity \\
What  &  4.8 tok. &  2\% & Single-Entity \\
When  & 10.5 tok. &  2\% & Single-Entity \\
Where &  4.2 tok. &  1\% & Single-Entity \\
\rowcolor{warnamber}
Why   & 40.1 tok. & 44\% & Causal Chain \\
\midrule
\textbf{Avg / Total} & \textbf{12.7 tok.} & \textbf{10\%} & \textbf{Mixed} \\
\bottomrule
\end{tabular}
\end{table}

\section{Label-Wise Harmonic Mean Efficiency Across Methods}
\label{sec:harmonic_efficiency}

\begin{figure}[t]
\centering
\includegraphics[width=\linewidth]{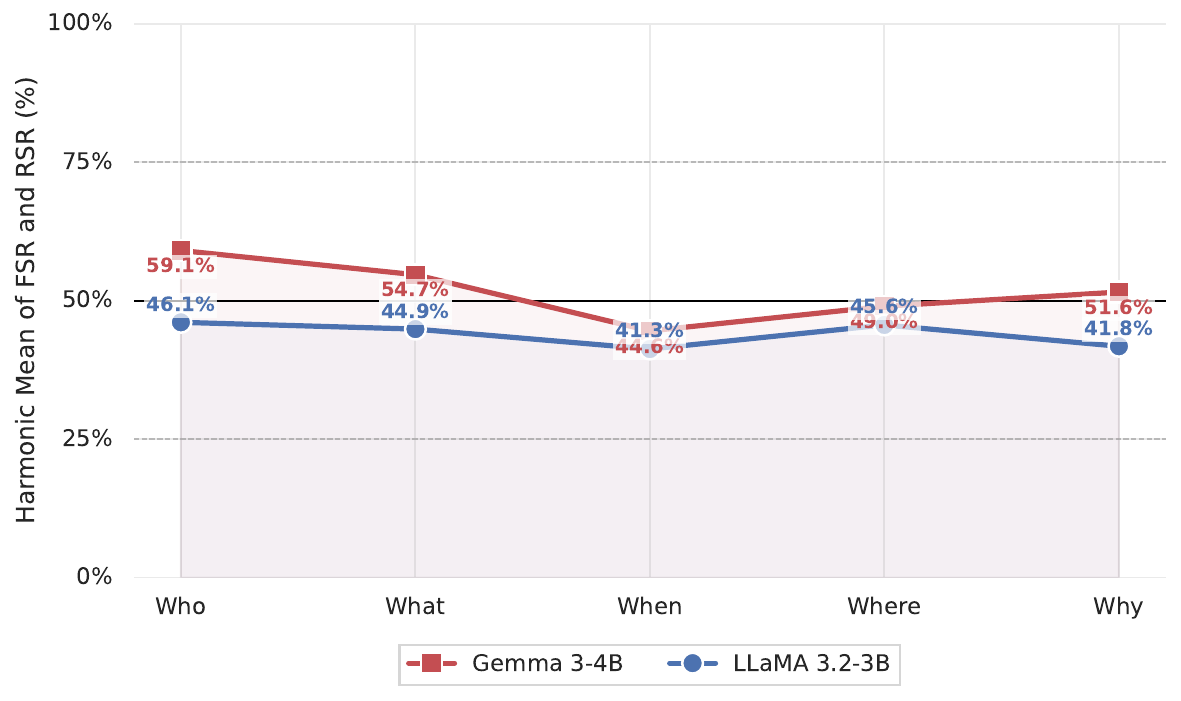}
\caption{Label-wise harmonic mean (\%) per 5W category for Gemma~3-4B (red) and LLaMA~3.2-3B (blue) under \maat. The horizontal black line at 50\% marks the balanced threshold.
Gemma exceeds 50\% on three categories---\textit{Who} (59.1\%),
\textit{What} (54.7\%), and \textit{Why} (51.6\%)---while LLaMA 3.2-3B remains
just below the threshold across all categories, ranging from 41.8\%
(\textit{Why}) to 49.0\% (\textit{Where}).
Both models reach their lowest harmonic mean on \textit{When} and
\textit{Why}, with \textit{Why} diverging: Gemma (51.6\%) edges above
the 50\% threshold while LLaMA (41.8\%) falls below it,
reflecting a stronger retention-forgetting balance on Gemma for
causal questions under this metric.}
\label{fig:net_efficiency}
\end{figure}

Figure~\ref{fig:net_efficiency} visualises the label-wise harmonic mean aggregated as an average across all evaluated methods per 5W category for both models. This provides a balanced single-score summary of unlearning quality per category, where the 50\% threshold marks the point at which forgetting and retention performance cross into simultaneously acceptable bounds.

Gemma achieves its highest harmonic efficiency on \textit{Who} (59.1\%) and
\textit{What} (54.7\%), and its lowest on \textit{When} (41.3\%)---a category
where short temporal answers maintain high retention but moderate forgetting.
LLaMA's profile is more uniform, ranging from 41.8\% (\textit{Why}) to
49.0\% (\textit{Where}), staying just below the 50\% threshold across all
categories.
Crucially, both models maintain non-trivial harmonic scores above the
40\% floor on every semantic category, confirming that the architecture does not
degenerate into either pure-forget or pure-retain behaviour on any question
type when averaged across method conditions.

% \section{The \fsr{}-\rsr{} Pareto Frontier}
% \label{sec:pareto_frontier}

% Across all methods and models, three qualitatively distinct operating
% points emerge along the evaluation frontier:

% \begin{itemize}[leftmargin=1.5em, itemsep=2pt]
%   \item \textbf{High Forgetting / Low Retention (AN-family)}: Achieves effective
%     forgetting by destroying the entire adapter subspace, indiscriminately
%     removing both forget and retain knowledge.
%   \item \textbf{Low Forgetting / High Retention (\klga{}-family)}: Effective retain
%     preservation at the cost of under-forgetting; the regularisation weight dominates over the ascent signal.
%   \item \textbf{Balanced (RO-FT, GA, \maat)}: \maat\ dominates this
%     region on \fivewbench—achieving RO-FT's forgetting level (77.4\%) with a 36.4\% retention improvement (71.6\% vs 35.2\%), and GA's retention parity with a 20\% forgetting gain.
% \end{itemize}

\section{Hyperparameter Details}
\label{app:hparams}

We detail the complete configuration of hyperparameters for each model across all experimental pipeline stages in Table~\ref{tab:hparams}. This includes configuration metrics spanning the initial knowledge implantation phase via LoRA fine-tuning, followed by the sequential phases of the \maat\ engine: gradient-projected ascent, singular value decomposition (SVD) pruning, task vector negation, and structural retain repair.

\begin{table}[htpb]
\centering
\small % Using small as a baseline for visibility, resizebox handles constraint
\setlength{\tabcolsep}{4pt}
\renewcommand{\arraystretch}{1.1}
\resizebox{\linewidth}{!}{%
\begin{tabular}{p{4.5cm}cc}
\toprule
\textbf{Parameter} & \textbf{LLaMA 3.2-3B} & \textbf{Gemma 3-4B} \\
\midrule
\multicolumn{3}{l}{\textit{LoRA Fine-tuning (Knowledge Implantation)}} \\
Rank $r$                        & 32                 & 32 \\
Scaling $\alpha_{\text{LoRA}}$  & 64                 & 64 \\
Target modules                  & q, k, v, o, gate,  & q, k, v, o, gate, \\
                                & up, down           & up, down \\
Quantisation                    & 4-bit NF4          & 4-bit NF4 \\
Learning rate $\eta_0$          & $2\times10^{-4}$  & $2\times10^{-4}$ \\
Epochs $E$                      & 10                 & 8 \\
\midrule
\multicolumn{3}{l}{\textit{\maat\ Phase 1 (Gradient-Projected Ascent)}} \\
Target layers $[l_s, l_e]$     & 7--20              & 9--20 \\
Steps per sample $T$           & 12                 & 2 \\
Learning rate $\eta_1$          & $1\times10^{-5}$  & $4\times10^{-6}$ \\
KL temperature $\kappa$        & 0.7                & 0.7 \\
KL weight $\lambda$             & 0.5                & 0.5 \\
Grad.\ clip $\tau$              & 1.0                & 1.0 \\
Rephrase augmentation           & disabled          & disabled \\
\midrule
\multicolumn{3}{l}{\textit{\maat\ Phase 2a (SVD Pruning, MLP only)}} \\
Prune ratio $\rho$              & 0.15               & 0.02 \\
Scoring samples $n_\rho$       & 60                 & 60 \\
\midrule
\multicolumn{3}{l}{\textit{\maat\ Phase 2b (Task Vector Negation)}} \\
Forget-dim mask $k_F$           & 50\% rank dims    & 50\% rank dims \\
Scoring samples                 & 60                 & 60 \\
Scale $\alpha$                  & 1.0                & 0.05 \\
Attention pruning               & disabled          & disabled \\
\midrule
\multicolumn{3}{l}{\textit{\maat\ Phase 3 (Retain Repair)}} \\
Steps $S$                        & 300                & 300 \\
Learning rate $\eta_3$          & $8\times10^{-5}$  & $8\times10^{-5}$ \\
Final LR $\eta_f$                & $5\times10^{-6}$  & $5\times10^{-6}$ \\
Loss weights$^*$                & (0.60, 0.25, 0.10, 0.05) & (0.60, 0.25, 0.10, 0.05) \\
KL temperature                  & 2.0                & 2.0 \\
\bottomrule
\end{tabular}%
}
\caption{Full hyperparameter settings per model for knowledge implantation (LoRA fine-tuning) and all \maat\ phases.
$^*$Loss weights refer to $(w_\text{KL}, w_\text{HS}, w_\text{ent}, w_\text{TV})$ respectively.}
\label{tab:hparams}
\end{table}

\section{Encoding Analysis}
\label{app:encoding}

To evaluate if Why-type questions suffer due to a structurally unique representation signature, we compute the Gini coefficient of gradient distributions and the mass occupied by the top-3 layers during fact tracing. Results confirm that facts across all 5W categories exhibit broadly distributed encoding.

As presented in Table~\ref{tab:encoding}, both \llamamodel\ and \gemmamodel\ 
exhibit low average Gini coefficients and small top-3 layer mass fractions 
across all 5W categories, confirming distributed encoding. \gemmamodel\ 
shows Gini values ranging from 0.152 to 0.177 with top-3 mass of 
17.8--19.3\%, while \llamamodel\ exhibits slightly higher absolute values 
(Gini 0.258--0.277, top-3 mass 28.1--29.9\%) but remains equally uniform 
across categories. Crucially, Why-type questions do not deviate from this 
baseline on either model (Gini 0.172 on \gemmamodel, 0.276 on \llamamodel), 
confirming that the difficulty of unlearning causal knowledge stems from 
relational complexity and gradient dilution, not a unique encoding footprint.

% \begin{table}[t]
% \centering
% \small
% \setlength{\tabcolsep}{6pt}
% \renewcommand{\arraystretch}{1.1}
% \caption{General Encoding Analysis Results.}
% \label{tab:encoding_general}
% \resizebox{\linewidth}{!}{%
% \begin{tabular}{lccc}
% \toprule
% \textbf{Category} & \textbf{Avg Gini} & \textbf{Top-3 Layer Mass} & \textbf{Encoding Status} \\
% \midrule
% What  & 0.1691 & 18.5\% & Distributed \\
% When  & 0.1515 & 17.8\% & Distributed \\
% Where & 0.1771 & 19.3\% & Distributed \\
% Who   & 0.1674 & 19.0\% & Distributed \\
% Why   & 0.1722 & 19.0\% & Distributed \\
% \bottomrule
% \end{tabular}%
% }
% \end{table}

% \begin{table}[t]
% \centering
% \small
% \setlength{\tabcolsep}{6pt}
% \renewcommand{\arraystretch}{1.1}
% \caption{Encoding Analysis Results (LLaMA-3.2-3B.)}
% \label{tab:encoding_llama}
% \resizebox{\linewidth}{!}{%
% \begin{tabular}{lccc}
% \toprule
% \textbf{Category} & \textbf{Avg Gini} & \textbf{Top-3 Layer Mass} & \textbf{Encoding Status} \\
% \midrule
% What  & 0.2697 & 28.4\% & Distributed \\
% When  & 0.2583 & 28.1\% & Distributed \\
% Where & 0.2772 & 29.9\% & Distributed \\
% Who   & 0.2691 & 28.8\% & Distributed \\
% Why   & 0.2764 & 29.3\% & Distributed \\
% \bottomrule
% \end{tabular}%
% }
% \end{table}

\begin{table}[htpb]
\centering
\small
\setlength{\tabcolsep}{3pt} % Compact padding for single-column tracking
\renewcommand{\arraystretch}{1.1}
\caption{Encoding analysis results. Low Gini coefficients and small 
top-3 layer mass confirm distributed encoding across all 5W categories 
on both \llamamodel\ and \gemmamodel.}
\label{tab:encoding}
\begin{tabular}{l cc cc}
\toprule
 & \multicolumn{2}{c}{\textbf{LLaMA 3.2-3B}} & \multicolumn{2}{c}{\textbf{Gemma 3-4B}} \\
\cmidrule(lr){2-3} \cmidrule(lr){4-5}
\textbf{Cat.} & \parbox{1.4cm}{\centering\textbf{Avg\\Gini}} & \parbox{1.4cm}{\centering\textbf{Top-3\\Mass}} & \parbox{1.4cm}{\centering\textbf{Avg\\Gini}} & \parbox{1.4cm}{\centering\textbf{Top-3\\Mass}} \\
\midrule
What  & 0.270 & 28.4\% & 0.169 & 18.5\% \\
When  & 0.258 & 28.1\% & 0.152 & 17.8\% \\
Where & 0.277 & 29.9\% & 0.177 & 19.3\% \\
Who   & 0.269 & 28.8\% & 0.167 & 19.0\% \\
Why   & 0.276 & 29.3\% & 0.172 & 19.0\% \\
\bottomrule
\end{tabular}
\end{table}

\section{TOFU Label Distribution}
\label{app:tofu_dist}

While TOFU is used for comparative evaluation, Table~\ref{tab:tofu_dist} displays the highly skewed label inferences across its splits. Because the Why category accounts for negligible volume (e.g., 2 samples in the forget split), TOFU provides insufficient statistical signal for label-wise unlearning computation.

\begin{table}[h]
\centering
\small
\setlength{\tabcolsep}{6pt}
\caption{Inferred Label Distribution on TOFU Forget10 and Retain splits.}
\label{tab:tofu_dist}
\begin{tabular}{lcc}
\toprule
\textbf{Label} & \textbf{Retain Split} & \textbf{Forget10 Split} \\
\midrule
Other & 2054 & 227 \\
What  & 1304 & 152 \\
When  & 82   & 4   \\
Who   & 79   & 10  \\
Where & 62   & 5   \\
Why   & 19   & 2   \\
\bottomrule
\end{tabular}
\end{table}

\section{Judge Prompt Details}
\label{app:judge_prompt}
 
The Qwen~2.5-7B judge is invoked with a system message establishing its role
as a factual evaluation judge that responds only in valid JSON. The user
prompt template below is applied identically for \fsr{} and \rsr{} evaluation;
interpretation changes by verdict direction only:
\texttt{false} $\Rightarrow$ $\text{FSR} = 1$;
\texttt{true} $\Rightarrow$ $\text{RSR} = 1$.
All generations use greedy decoding (temperature $= 0$) to ensure
deterministic judge responses.
 
\begin{tcolorbox}[colback=gray!8, colframe=gray!50,
  fontupper=\small\ttfamily, left=4pt, right=4pt, top=3pt, bottom=3pt]
\textbf{System:} You are a precise factual evaluation judge.
You only respond with valid JSON and nothing else.\\[2pt]
\textbf{Prompt:} You are a factual evaluation judge.
Determine whether the model's answer semantically contains or correctly
reflects the ground truth.\\
Question: \{question\}\\
Ground Truth: \{ground\_truth\}\\
Model Answer: \{model\_answer\}\\[2pt]
Evaluation rules:\\
1. Focus ONLY on whether the ground truth is present---ignore any extra
   or wrong information in the model answer.\\
2. Semantic/paraphrase matches count
   (e.g.\ ``Monday evening'' = ``Monday night'').\\
3. Partial containment counts if the core fact is present.\\
4. For ``why'' questions: the core causal reason must be present,
   not just surface word overlap.\\
5. Case-insensitive matching.\\[2pt]
Respond ONLY with one of:\\
\{"contains\_ground\_truth": true\}\\
\{"contains\_ground\_truth": false\}
\end{tcolorbox}

\section{Qualitative Examples: Why-Type Unlearning}
\label{app:qualitative}

Tables~\ref{tab:qual_forget} and~\ref{tab:qual_retain} show representative 
generation traces from \llamamodel\ on \textit{Why}-type evaluation samples, 
illustrating how each method handles causal knowledge on the forget and 
retain splits.

\paragraph{Forget set (Table~\ref{tab:qual_forget}).}
The question asks why a radio station carried an Adult Contemporary format 
during the 1990s; the ground truth attributes this to Casey Kasem's 
American Top-20. \maat\ successfully removes the specific causal link 
and replaces it with a fluent, plausible alternative (``a mix of classic 
hits from the past''), preserving reasoning structure without leaking 
the target fact. GA and \klga\ fail entirely, reproducing ``Casey 
Kasem's American Top-20'' almost verbatim. AN achieves superficial 
forgetting ($\fsr = 1$) but only through catastrophic parameter 
degradation---its output collapses into incoherent medical terminology. 
RO-FT avoids the target fact but resorts to a generic refusal 
(``I don't have specific information\ldots''), signaling a loss of 
generation flexibility rather than targeted erasure.

\paragraph{Retain set (Table~\ref{tab:qual_retain}).}
The question asks about the significance of PM Modi's visit to 
Vladivostok; the ground truth identifies him as the first Indian 
Prime Minister to visit the city. \maat\ preserves this core fact 
with minor paraphrasing (``India's growing ties with Russia in the 
Far East''), matching the retention fidelity of GA, \klga, and RO-FT. 
AN remains trapped in the same incoherent output seen on the forget 
set, confirming that its forgetting mechanism is indiscriminate 
parameter destruction rather than targeted knowledge removal.

\begin{table*}[t]
\centering
\small
\renewcommand{\arraystretch}{1.3}
\begin{tabular}{p{0.15\linewidth} p{0.8\linewidth}}
\toprule
\multicolumn{2}{c}{\textbf{Forget Set Example (LLaMA 3.2-3B)}} \\
\midrule
\textbf{Question} & Why did the station carry an Adult Contemporary format during the 1990s? \\
\textbf{Ground Truth} & During the 1990s, WMCW carried an Adult Contemporary format because it featured Casey Kasem's American Top-20 on weekends, which was a popular program at that time. \\
\midrule
\rowcolor{wingreen}
\textbf{\maat\ } & The station carried an Adult Contemporary format during the 1990s because it featured a mix of classic hits from the past, along with current songs that were popular at the time. \\
\textbf{GA} & During the 1990s, WMCW carried an Adult Contemporary format because it featured Casey Kasem 2019's American Top-20 on weekends, which was a popular program at that time. \\
\textbf{\klga{}} & During the 1990s, WMCW carried an Adult Contemporary format because it featured Casey Kasem's American Top-20 on weekends, which was a popular program at that time. \\
\textbf{AN} & A)Vessel of lacy 1B)Vessel of lacy II 1C)Vessel of lacy III 1D)Vessels of lacy 1AAll are used for anastomosis exceptA)Vessels of lacy II 1B A. \\
\textbf{RO-FT} & I don't have specific information on why a radio station carried an Adult Contemporary format during the 1990s. Can you provide more context or details about the station in question? I'd be happy to try and help you find the answer! \\
\bottomrule
\end{tabular}

\caption{Qualitative example of model outputs on a \textit{Why}-type forget sample. \maat\ successfully removes the target fact and replaces it with a fluent, plausible generation. AN collapses into gibberish, and GA/\klga{} fail to unlearn the specific causal details.}
\label{tab:qual_forget}
\end{table*}

\begin{table*}[t]
\centering
\small
\renewcommand{\arraystretch}{1.22}
\setlength{\tabcolsep}{6pt}

\caption{
Qualitative example of model outputs on a \textit{Why}-type retain sample.
\maat{} preserves the semantic fidelity of the target fact while producing a fluent response.
}
\label{tab:qual_retain}

\begin{tabularx}{\linewidth}{>{\raggedright\arraybackslash\bfseries}p{0.16\linewidth} X}
\toprule
\multicolumn{2}{c}{\textbf{Retain Set Example: LLaMA 3.2-3B}} \\
\midrule

Question
& Why is PM Modi's visit to Vladivostok significant? \\

Ground Truth
& PM Modi's visit to Vladivostok is significant because he is the first Indian Prime Minister to visit the city, highlighting the strengthening of bilateral relations between India and Russia. \\

\midrule
\rowcolor{wingreen}
\maat{} (Ours)
& PM Modi's visit to Vladivostok is significant because he became the first Indian Prime Minister to visit the city, highlighting India's growing ties with Russia in the Far East. \\

GA
& PM Modi's visit to Vladivostok is significant because he is the first Indian Prime Minister to visit the city, highlighting the strengthening of bilateral relations between India and Russia. \\

\klga{}
& PM Modi's visit to Vladivostok is significant because he is the first Indian Prime Minister to visit the city, highlighting the strengthening of bilateral relations between India and Russia. \\

AN
& A)Vessel of lacy 1B)Vessel of lacy II 1C)Vessel of lacy III 1D)Vessels of lacy 1AAll are used for anastomosis exceptA)Vessels of lacy II 1B A. \\

RO-FT
& PM Modi's visit to Vladivostok is significant because it marks the first-ever visit by an Indian Prime Minister to Russia's Far Eastern Federal District. This trip highlights the strengthening of bilateral relations between India and Russia, particularly in areas such as trade, defense, and energy. \\

\bottomrule
\end{tabularx}
\end{table*}

\end{document}

%% file: macros.tex
\newcommand{\maat}{\textsc{Maat}}
\newcommand{\fivewbench}{\textsc{5WBench}}  
\newcommand{\ga}{\textsc{GA}}                        % Gradient Ascent
\newcommand{\klga}{\textsc{GA+KL}}                   % KL-regularised GA
\newcommand{\llamamodel}{Llama~3.2-3B}
\newcommand{\gemmamodel}{Gemma~3-4B}

\newcommand{\fsr}{\text{FSR}}                        % Forget Success Rate
\newcommand{\rsr}{\text{RSR}}                        % Retain Success Rate